\documentclass[10pt,twocolumn,letterpaper]{article}

\usepackage[pagenumbers]{cvpr} 

\usepackage[dvipsnames]{xcolor}

\definecolor{cvprblue}{rgb}{0.21,0.49,0.74}
\usepackage[pagebackref,breaklinks,colorlinks,citecolor=cvprblue]{hyperref}
\usepackage{booktabs}
\usepackage{multirow}
\usepackage{multicol}
\usepackage{pifont}
\usepackage{stfloats}

\def\paperID{13678} 
\def\confName{CVPR}
\def\confYear{2024}

\title{Autoregressive Queries for Adaptive Tracking with Spatio-Temporal Transformers}

\author{Jinxia Xie\textsuperscript{1,2,3}, Bineng Zhong\textsuperscript{1,2,3}\thanks{Corresponding Author}, Zhiyi Mo\textsuperscript{3}, Shengping Zhang\textsuperscript{4}, Liangtao Shi\textsuperscript{1}, Shuxiang Song\textsuperscript{1}, Rongrong Ji\textsuperscript{5}\\
\textsuperscript{1}Key Laboratory of Education Blockchain and Intelligent Technology Ministry of Education\\
\textsuperscript{2}Guangxi Key Lab of Multi-Source Information Mining \& Security\\
\textsuperscript{1,2}Guangxi Normal University, Guilin 541004, China\\
\textsuperscript{3}Guangxi Key Laboratory of Machine Vision and Intelligent Control, Wuzhou University\\
\textsuperscript{4}School of Computer Science and Technology, Harbin Institute of Technology\\
\textsuperscript{5}Media Analytics and Computing Lab, School of Informatics, Xiamen University\\
{\tt\small xie\_jx@stu.gxnu.edu.cn, bnzhong@gxnu.edu.cn, zhiyim@gxuwz.edu.cn, s.zhang@hit.edu.cn}
\\
{\tt\small slt@stu.gxnu.edu.cn, songshuxiang@mailbox.gxnu.edu.cn,  rrji@xmu.edu.cn } %
}

\def\mytracker{AQATrack}

\begin{document}
\maketitle

\begin{abstract}
The rich spatio-temporal information is crucial to capture the complicated target appearance variations in visual tracking. 
However, most top-performing tracking algorithms rely on many hand-crafted components for spatio-temporal information aggregation. 
Consequently, the spatio-temporal information is far away from being fully explored. To alleviate this issue, we propose an adaptive tracker with spatio-temporal transformers (named {\mytracker}), which adopts simple autoregressive queries to effectively learn spatio-temporal information without many hand-designed components. Firstly, we introduce a set of learnable and autoregressive queries to capture the instantaneous target appearance changes in a sliding window fashion. Then, we design a novel attention mechanism for the interaction of existing queries to generate a new query in current frame. 
Finally, based on the initial target template and learnt autoregressive queries, a spatio-temporal information fusion module (STM) is designed for spatio-temporal formation aggregation to locate a target object. 
Benefiting from the STM, we can effectively combine the static appearance and instantaneous changes to guide robust tracking. 
Extensive experiments show that our method significantly improves the tracker's performance on six popular tracking benchmarks: LaSOT, LaSOT$_{ext} $, TrackingNet, GOT-10k, TNL2K, and UAV123.
Code and models will be \href{https://github.com/orgs/GXNU-ZhongLab}{here}.

\begin{figure}[t]
    \centering
    \includegraphics[height=4.8cm]{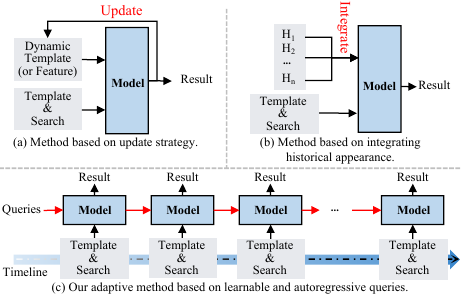}
    \caption{The comparison of three different tracking paradigms. (a) The trackers based on an updating strategy to update a dynamic template or feature. (b) The trackers based on integrating historical appearance. H represents historical appearance. (c) The proposed adaptive tracker with learnable and autoregressive queries.}
    \label{fig:1}
\end{figure}
  
\end{abstract}   
\section{Introduction}
Visual object tracking(VOT) is a fundamental task in computer vision, which aims to estimate the position and shape of an arbitrary target in video sequences given its initial status. 
It has a wide range of applications in fields including robotic vision, video surveillance \cite{surveillance0, surveillance1}, autonomous driving \cite{autodriving0, autodriving1}, and various other domains. 
However, the tracking is often influenced by many factors, including camera movement, self-deformation, and external environment (occlusion or distractions from similar objects). 
And the target appearance always changing. 

Due to the aforementioned challenges, mainstream tracking algorithms \cite{transt, aiatrack, simtrack, ostrack, SBT, sparseTT} are difficult to effectively discriminate targets based on the static appearance (initial template).
Therefore few models \cite{seqformer, stark, mixformer, KYS, SwinTrack, CTTrack} explore spatio-temporal information to capture the appearance changes to improve their discriminative ability.
These methods mine the spatio-temporal information in two main manners.
The first one relies on an update strategy to update a new target appearance, as shown in \cref{fig:1}(a). 
Some trackers \cite{seqtrack, CTTrack, mixformer} update a dynamic template to get a new target appearance.
In addition, some trackers \cite{KYS, SwinTrack} get the new target appearance by updating a state feature or motion feature.
These methods based on an update strategy use a confidence score to update the template or feature in an interval.
Even though these methods have achieved success, they require a manual design of update strategies and introduce hyperparameters (e.g., intervals, thresholds).
The second approach to mining the spatio-temporal information is to integrate the historical target appearance, as illustrated in \cref{fig:1}(b).
These methods \cite{ToMP, VideoTrack1, STMTrack, updateNet} integrate historical appearance through some operation, including feature concatenation \cite{stark, VideoTrack1},  weighted sum \cite{updateNet}, and memory networks \cite{STMTrack}. 
Although these methods achieve a competitive performance, they require more computational resources and are more likely to lead to error accumulation.
%

To avoid the above issues, we propose an adaptive tracker (named {\mytracker}) with spatio-temporal transformers, which adopts simple autoregressive queries to effectively learn spatio-temporal information without cumbersome and customized components, as demonstrated in \cref{fig:1}(c). 
Firstly, we use the HiViT \cite{hivit} as the encoder, whose task is to learn outstanding spatial features of the target.
Secondly, we design a decoder to mine and propagate spatio-temporal information across continuous frames.
We introduce a set of learnable and autoregressive \textit{target queries} to capture the instantaneous target appearance changes in a sliding window fashion.
And \textit{temporal attention} mechanism is used for the interaction of existing queries to generate a new query in the current frame.
Finally, a spatio-temporal information fusion module (STM) is designed for spatio-temporal information aggregation to locate a target object without any hyperparameters. 
Benefiting from the STM, we can effectively combine the static and instantaneous target appearance changes to guide robust tracking.
Detailed experiment shows that our method can effectively capture the target state changes and motion trends.
Our main contributions are summarized as follows:
\begin{itemize}
    \item To fully explore the spatio-temporal information, we propose an adaptive tracker to capture instantaneous appearance changes without any hand-designed components.
    \item In the proposed tracker, we introduce a set of learnable and autoregressive queries to capture the instantaneous target appearance changes in a sliding window fashion. A spatio-temporal information fusion module is designed to combine static appearance and instantaneous changes.
    \item Extensive experimental results demonstrate that our tracker achieves SOTA performance on six challenging benchmarks. In particular, \mytracker-256 and \mytracker-384 achieves 71.4\% and 72.7\% AUC score on long-term benchmarks LaSOT \cite{lasot}, respectively.
\end{itemize}

\section{Related Work}
\label{sec:Related work}
\textbf{Visual object tracking based on spatial features.}
Most trackers perform well by introducing a backbone with powerful spatial feature extraction capabilities from detection tasks or natural language processing (NLP) tasks.
SiamFC \cite{SiamFC} designed a siamese network framework using AlexNet \cite{alexnet} as the backbone network to extract features from template and search, which achieved good performance in speed and accuracy.
Some trackers \cite{transt, stark} used ResNet \cite{resnet1} as a backbone and achieved excellent performance. 
In recent years, the transformer-based algorithms introduced to target recognition demonstrated astonishing global modeling capabilities.
Therefore, visual object tracking also began to use transformers 
\cite{transformer}. Initially, some trackers utilized an attention mechanism or transformer for feature extraction or fusion in tracking. 
Such as TransT \cite{transt} designed two modules based on the attention mechanism for feature interaction. STARK \cite{stark} uses the transformer structure as a fusion module. Later, due to the unstoppable charm of the transformer, some trackers \cite{SwinTrack, mixformer}  took the transformer as a backbone. 
In addition, some researchers proposed full transformer-based trackers \cite{ostrack, SBT, mixformer} which join feature extraction and fusion, greatly improving the performance.

\textbf{Tracking combining spatio-temporal information.}
Spatio-temporal information is vital for the model to capture the target state changes and motion trends.
Thus many mainstream studies \cite{seqtrack, VideoTrack1, ARTrack, CTTrack, mixformer} explored spatio-temporal information in visual object tracking. 
One usually used method is updating appearance representation.
Many works \cite{seqtrack, CTTrack, mixformer} adopted a dynamic template to update the target appearance to capture changes, thus making the matching between template and search images more accurate.
A few studies \cite{KYS, SwinTrack} focused on learning a feature to describe the target's previous state or motion information.
Another common method for exploring spatio-temporal information is integrating historical appearance. 
STMTrack \cite{STMTrack} proposes a space-time memory network to make full use of historical information.
UpdateNet \cite{updateNet} takes into account a set of historical appearances to estimate the optimal template for the next frame.
However, most of these methods require the design with some artificial rules, and the spatio-temporal information is far from being fully explored.
Recently, TCTrack\cite{TCTrack}/TCTrack++\cite{TCTrack++} are proposed for temporal contexts in aerial tracking.
They exploit spatio-temporal information on two levels: the extraction of features and the refinement of similarity maps.
ARTrack\cite{ARTrack} autoregressively predicts the current coordinates based on historical coordinates.
\begin{figure*}[t] 
    \centering
    \includegraphics[height=5cm]{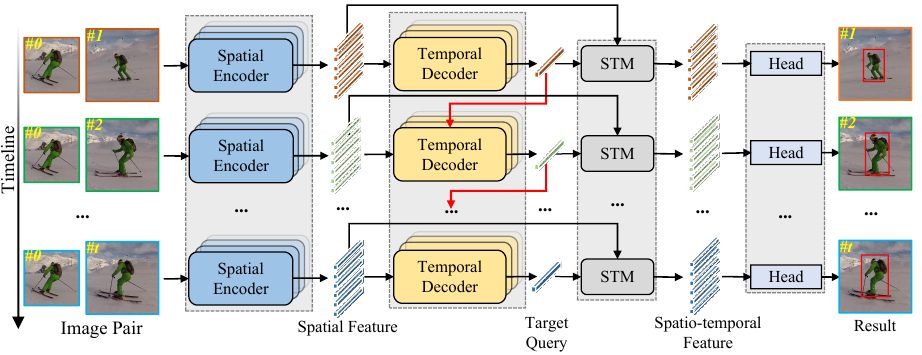}
    \caption{Overview of our framework. It mainly consists of four components, i.e., a spatial encoder for spatial features, a temporal decoder for learning an autoregressive target query that incorporates temporal information(with {\color{red}red} arrows), a spatio-temporal feature fusion module(STM) designed for a spatio-temporal feature, and a prediction head. }
    \label{fig:2}
\end{figure*}

\textbf{The utilization of query.}
DETR \cite{DETR} introduced the concept of \textit{query} to detect different objects. Since then, many fields explored the usage of queries, i.e. video instance segmentation (VIS), multiple object tracking (MOTR), and Video object detection (VOD). 
Some algorithms \cite{motr1, TrackFormer, seqformer, VisTR} use queries to help recognize the target, which collects information about the changing state of the object in the video clip.
In the field of video segmentation, VisTR \cite{VisTR} adapts transformer \cite{transformer} to VIS \cite{VIS1} and uses instance queries to obtain instance sequences from video clips.
In the realm of multiple object tracking, MOTR \cite{motr1} introduced \textit{track query} to model the tracked instances in the entire video, transferring and updating it frame-by-frame for iterative predictions over time.
Inspired by the usage of queries in so many excellent algorithms, we propose a decoder based on target queries to explore spatio-temporal information within video clips. To the best of our knowledge, we are the first to introduce queries in an autoregressive manner in single object tracking.

\section{Method}
In this section, we provide a detailed explanation of our proposed tracker. We start by giving a concise overview of our spatio-temporal tracking framework. Next, we delve into the specific components of our model, including the spatial encoder, temporal decoder with queries and temporal attention, as well as the spatio-temporal information fusion modules (STM). Finally, we introduce the head network and loss function used in the tracker.

\subsection{Overview}
As shown in \cref{fig:2}, our proposed tracker primarily consists of a spatial encoder, a temporal decoder, and a spatio-temporal feature fusion operation module (STM).
A pair of images are input into the encoder, including
a template image $\mathbf{Z}\in\mathbb{R}^{H_{z}\times W_{z}\times3}$
and a search region image $\mathbf{X}\in\mathbb{R}^{H_{x}\times W_{x}\times3}$.
The spatial encoder uses a hierarchical downsampling method for image processing and then learns outstanding representation features through attention mechanisms.
The temporal decoder takes two inputs: 
the first one is a spatial feature from the spatial encoder, and the second one is some learnable and autoregressive queries. Its task is to learn the target appearance changes to better guide the expression of spatial features.
In addition, we employed a spatio-temporal information fusion module (STM) to combine the static appearance and instantaneous 
appearance changes.
At last, the output features of STM will be used for result prediction. We refer to the previous work and employ a center head network to predict the result.

\subsection{Spatial Encoder}
    Many tracking algorithms \cite{ostrack, simtrack, GRM} use a ViT \cite{vit} as the backbone, and its patch size for patch embedding is 16×16. However, we believe that conducting a large downsampling at once will weaken the correlation between different patches.
    So we use a gradual downsampling network as our spatial encoder, which is with a 4×4 patch size for patch embedding. A total of eight multi-layer perceptron (MLP) layers and two merging layers were used to achieve the goal of gradually downsampling.
    After the above operation, the obtained search tokens and template tokens are $f_{z}\in\mathbb{R}^{N_{z}\times D}$ and $f_{x}\in\mathbb{R}^{N_{x}\times D}$, respectively.
    Here, $N_{z}=H_{z}W_{z}/16^{2}$, $N_{x}=H_{x}W_{x}/16^{2}$, $D=512$. 
    Next, the template token and search token will be concatenated and fed into the N-layer encoder for spatial feature learning.
    The operation in our encoder can be described as the following equations:
    \begin{equation}
        \begin{split}
            &f_{zx}^0 = Concat(f_{z}, f_{x}), \\
            &f_{zx}^n = Encoder(f_{zx}^{n-1}), n=1...N, \\
            &f_{zx}^N = LN(f_{zx}^N).
            \label{eq:enc}
        \end{split}
    \end{equation}
    Refer to HiViT \cite{hivit} for a more detailed design of our encoder.

\subsection{Temporal Decoder}
    If only spatial features are used for tracking and prediction, it cannot effectively cope with challenges such as motion or interference from similar objects. 
    Therefore, it is necessary to introduce temporal information to guide the expression of spatial features. 
    \begin{figure}
    \centering
    \includegraphics[width=0.8\linewidth]{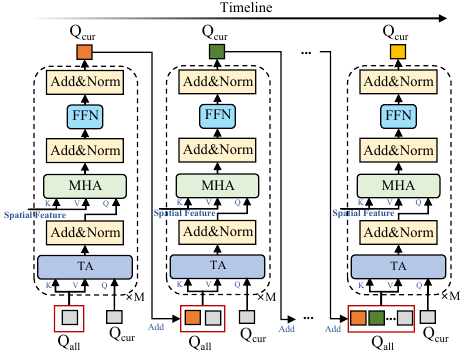}
    \caption{The structure of the temporal decoder is equipped with target query and temporal attention. Here FFN, MHA, and TA are feedforward neural networks, multi-head attention, and temporal attention, respectively. And $Q$ represents the target query.}
    \label{fig:3-}
\end{figure}
    The number of our temporal encoder layers is $M$, and each layer mainly includes the following parts: a temporal attention (TA) layer, a multi-head attention (MHA) layer, and a feedforward neural network (FFN) layer.
    There are two inputs to the temporal encoder, one is the spatial feature $f_{zx}^N$ from the spatial encoder, and the other is temporal queries.

    Here, let's review the multi-head attention mechanism, whose formula is as follows:
    \begin{equation}
        \begin{split}
            &\operatorname{MultiHead}(\mathbf{Q},\mathbf{K},\mathbf{V})=\operatorname{Concat}(\mathbf{H}_1,...,\mathbf{H}_{n_h})\mathbf{W}^O,\\ 
            &\mathbf{H}_{i}=\mathrm{Attention}(\mathbf{QW}_{i}^{Q},\mathbf{KW}_{i}^{K},\mathbf{VW}_{i}^{V}),\\
            &\mathrm{Attention}(\mathbf{q},\mathbf{k},\mathbf{v})=\mathrm{Softmax}(\frac{\mathbf{q}\mathbf{k}^\top}{\sqrt{d_k}})\mathbf{v}. \\
        \end{split}  
        \label{eq:attn}
    \end{equation}
    The following are learnable parameters:
    $\mathbf{W}_i^Q\quad\in\quad\mathbb{R}^{d_m\times d_k}$, 
    $\mathbf{W}_i^K\quad\in\quad\mathbb{R}^{d_m\times d_k}$, 
    $\mathbf{W}_i^V\quad\in\quad\mathbb{R}^{d_m\times d_v}$,    
    and $\mathbf{W}^O\quad\in\quad\mathbb{R}^{n_h\times d_m}$ are learnable parameter. 
    In {\mytracker}, we employ multi-head attention with 8 heads, i.e., $n_h=8$, $d_m=512$, and $d_k=d_v=d_m/8=64$.

    \textbf{Target query.}
    Inspired by works \cite{TrackFormer, motr1, TransTrack} in multiple object tracking (MOT) tasks, we introduced queries to capture the spatio-temporal information, called \textit{target query}. 
    To our knowledge, we are the first to introduce autoregressive queries for mining spatio-temporal information in the single object tracking (SOT) task.
    As shown in the \cref{fig:3-}, there are two types of query in our decoder: one is passed down from the previous frames called $Q_{pre}$, and the other is the query to be learned from the current frame called $Q_{cur}$. 
    $Q_{cur}$ can describe the state of the target in the current frame and integrate spatio-temporal information.
    The decoder input consists of $Q_{all}$ and $Q_{cur}$, which can be describe as folows:
    \begin{equation}
        \begin{split}
            &Q_{all} = Concat(Q_{pre},Q_{cur}),\\
            &Q_{pre}=\begin{cases}
            Concat(Q_{1},...,Q_{t-1}), &t<m\\
                Concat(Q_{t-m+1},...,Q_{t-1}),&t\geq m 
            \end{cases}
            \label{eq:Q}
        \end{split}
    \end{equation}
    where $m$ is the length of spatio-temporal information.
    The $Q_{cur}$ after the temporal decoder is propagated to the next frame of the video clip as one of the $Q_{pre}$.
    Thus, the temporal decoder learns spatio-temporal information in the form of sliding windows.

    \textbf{Temporal attention (TA).}
    Self-attention gives equal attention to all queries, resulting in an inability to achieve pure temporal information from the $Q_{pre}$.
    In order to capture elaborate target state changes and motion trends, we design temporal attention for the interaction of existing queries to generate a new query in current frame.
    As shown in \cref{fig:3-}, $Q$ is generated by $Q_{cur}$, $K$ and $V$ generated by $Q_{all}$. 
    The calculation of temporal attention is the same as the \cref{eq:attn}.
    
\subsection{Spatio-temporal Information Fusion Module}
    We used a simple and effective operation to fuse the spatial features and temporal information without introducing any parameters. Specifically, we first use a dot product to calculate the similarity $S\in\mathbb{R}^{N_{x} \times N_{temporal}}$ between search spatial features $f_{x}^N\in\mathbb{R}^{N_{x} \times D}$ and temporal information $f_{temporal}\in\mathbb{R}^{N_{temporal} \times D}$, where $D=512$. The parameter-free operation can be described as the following equation: 
    \begin{equation}
        \begin{split}
            S = f_{x}^N\odot f_{temporal}^\top, 
            \label{eq:fusion}
        \end{split}
    \end{equation}
    where $\odot$ means \textit{Dot product}.
    The similar scores $S$ highlight the location where the target may be located. Then use it to enhance the expression of spatial feature $f_{x}^N$ with \textit{element-wise product}. 
\subsection{Head and Loss}
We use a center-based head for predicting the centroid position and scale of the target. 
The outputs of the prediction head are the classification score map $P\in[0,1]^{\frac{H_x}P\times\frac{W_x}P}$, the size of the bounding box $B\in[0,1]^{2\times\frac{H_x}{P}\times\frac{W_x}{P}}$, and the offset size $O\in[0,1)^{2\times\frac{H_x}P\times\frac{W_x}P}$. 
The location with the highest classification score is taken as the target's location, and the final tracking result is calculated by combining the offset size and the bounding box size.
We use focal \cite{focalloss} loss and GIoU \cite{giouloss} loss in the prediction head network, the total loss $L$ calculation can be described as:
\begin{equation}
    L=L_{cls}+\lambda_{iou}L_{iou}+\lambda_{L_1}L_1,\\
    \label{eq:loss}
\end{equation}
where $\lambda_{iou}=2$ and $\lambda_{L1}=5$ are the regularization parameters.

\section{Experiments}
\begin{figure}[t]
    \centering
    \includegraphics[height=6cm]{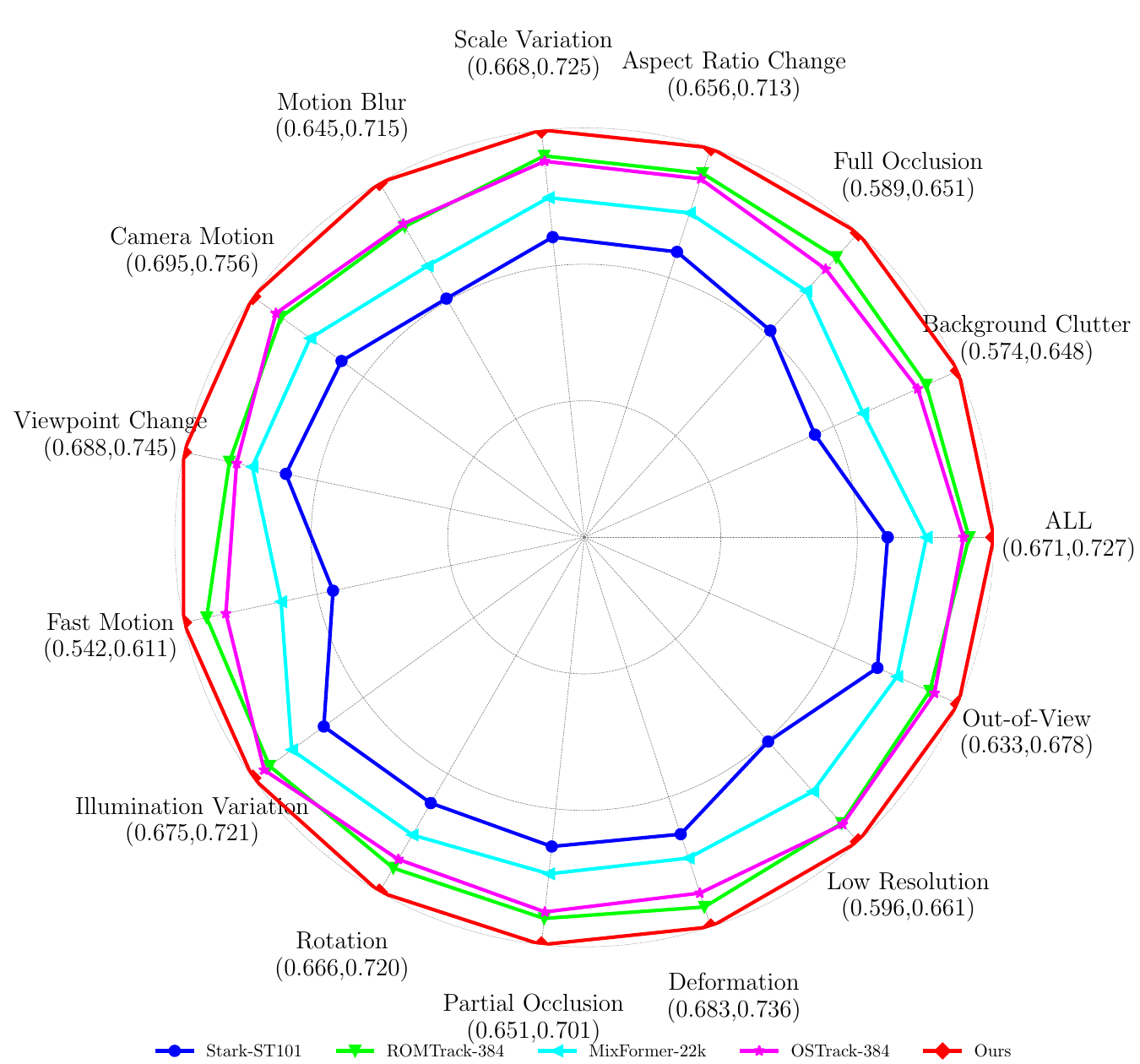}
    \caption{AUC scores of difference attributes on LaSOT\cite{lasot}. Best viewed in color.}
    \label{fig:zhizhuwang}
\end{figure} 

\subsection{Implementation Details}
Our algorithm uses Pytorch 1.9.0 in Python 3.8. Our tracker was trained and tested on 4 NVIDIA v100 GPUs.

\begin{table}[t]
    \caption{ Comparison of model parameters, FLOPs, and inference.}
    \centering
    \resizebox{\linewidth}{!}{
        \fontsize{9}{11}\selectfont
        \begin{tabular}{c|ccccc}
        \toprule
         Model & Device & Speed(FPS) & MACs(G) & Params(M)  & AUC(\%) \\
         \midrule
         SeqTrack-256\cite{seqtrack} & 2080Ti & 40 & 66  & 89 & 69.9  \\
         STARK\cite{stark}& Tesla v100 & 31.7 & 18.5 & 43.4 & 67.1  \\
         \midrule
         {\mytracker}-256(ours) & Tesla v100 & 67.6 & 25.8 & 72 & 71.4  \\
         {\mytracker}-384(ours) & Tesla v100 & 44.2 & 58.3  & 72 & 72.7  \\
        \bottomrule 
        \end{tabular}
    }
    
    \label{tab:fps}
\end{table} 
\textbf{Model variants.} 
We present two variants of {\mytracker} with different configurations as follows: 
\begin{itemize}[leftmargin=0.9cm]
    \item \textit{\mytracker-256.} Template size:[128×128]; Search region size: [256×256]; 
    \item \textit{\mytracker-384.} Template size:[192×192]; Search region size:[384×384].
\end{itemize}
{\mytracker} mainly includes three parts: spatial encoder, temporal decoder, and spatio-temporal information fusion module (STM). We use HiViT-Base \cite{hivit} as the spatial encoder with a layer $N$ of 20. In terms of the decoder, the number of layers $M$ is 3, the hidden size is 256, the number of attention heads is 8, and the hidden size of the feed forward network (FFN) is 512. 
The number of temporal queries $Q_{all}$ is four, of which three are previous temporal queries $Q_{pre}$ and one is the current target query $Q_{cur}$. 

\textbf{Training strategy.} Following traditional protocols, we trained our model on four datasets, namely: LaSOT \cite{lasot}, COCO \cite{coco}, TrackingNet \cite{trackingnet}, and GOT-10k \cite{got10k} (remove 1,000 videos as \cite{stark}).
We test the GOT-10k with its training split to follow the protocol described in \cite{got10k}. 
We use typical commonly used data augmentation methods, including horizontal flipping and brightness jittering.
We train {\mytracker} with AdamW \cite{adamw} optimizer, set the weight decay to $10^{-4}$, the initial learning rate for the backbone to 4 × $10^{-5}$, and other parameters to 4 × $10^{-4}$.
We trained the model for a total of 150 epochs with 60k image pairs, and the learning rate was decayed to 4 × $10^{-5}$ at the 120th epoch.
For GOT-10k, we trained the model for a total of 100 epochs, and the learning rate decayed at the 80th epoch.
To learn continuous spatio-temporal information, we used video-level sampling strategies in training. Specifically, we sample $n$ video sequences each containing $m$ template-search pairs (with the same template).
So, the total batch size in each iteration is $n*m$.
For the \mytracker-256, we set $n$ and $m$ to 4,8, respectively. For \mytracker-384, due to the limitations of GPU memory, $n$ and $m$ are 4 and 4 respectively.

\begin{figure}[t]
    \centering
    \includegraphics[width=\linewidth]{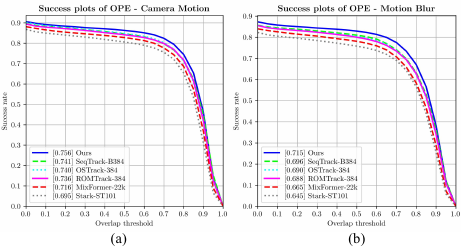}
    \caption{Success plots of one-pass evaluation (OPE) about camera motion and motion blur challenges on LaSOT\cite{lasot}. Best viewed in color and zooming in.}
    \label{fig:motion}
\end{figure} 
\textbf{Inference.}
We followed the common practice \cite{ostrack, transt, Ocean} and utilized the Hamming window to incorporate positional priors.
We use a set of target queries provided by the previous frames in the inference process to mine instantaneous appearance changes.
As demonstrated in the \cref{tab:fps}, we compared inference speed, MAC, and Params with state-of-the-art trackers. Our {\mytracker} can run in real-time at more than $65fps$ which is more than twice as far as STARK \cite{stark}.

\begin{table*}[t]
\caption{ Performance comparisons with state-of-the-art trackers on the test set of LaSOT, $\rm LaSOT_{ext}$ , GOT-10K, TNL2K and UAV123. We add a symbol * over GOT-10k to indicate that the corresponding models are only trained with the GOT-10k training set. The top two results are highlighted with {\color{red}red} and {\color{blue}blue} fonts, respectively.}
\centering
\resizebox{\textwidth}{!}{
    \fontsize{9}{10.8}\selectfont
    \begin{tabular}{r|c|ccc|ccc|ccc|cc|c}
    \toprule
    \multicolumn{1}{c|}{\multirow{2}{*}{Method} }
    & \multicolumn{1}{c|}{\multirow{2}{*}{Source}} 
    & \multicolumn{3}{c|}{LaSOT\cite{lasot}} 
    & \multicolumn{3}{c|}{$\rm LaSOT_{ext}$} 
    & \multicolumn{3}{c|}{GOT-10K$^*$} 
    & \multicolumn{2}{c|}{TNL2K}
    & \multicolumn{1}{c}{UAV123}\\ 
    \cline{3-14}
                                            && AUC(\%) & $\rm P_{norm}$(\%) & P(\%)     & AUC(\%) & $\rm P_{norm}$(\%) & P(\%)      & AO(\%) & $\rm SR_{0.5}$(\%) & $\rm SR_{0.75}$(\%)     & AUC(\%) & P(\%) & AUC(\%) \\
    \midrule
    {\mytracker}-256 & Ours                      & \textbf{\color{red}{71.4}} & \textbf{\color{red}{81.9}} & \textbf{\color{red}{78.6}}    & \textbf{\color{red}{51.2}} & \textbf{\color{red}{62.2}} & \textbf{\color{red}{58.9}}    & \textbf{\color{blue}{73.8}} & \textbf{\color{blue}{83.2}} & \textbf{\color{red}{72.1}}    &\textbf{\color{red}{57.8}}   &\textbf{\color{red}{59.4}}  &\textbf{\color{red}{70.7}}\\
    \midrule
    F-BDMTrack-256\cite{F-BDMTrack}&ICCV23  & 69.9 & 79.4 & 75.8    & 47.9 & 57.9 & 54.0    & 72.7 & 82.0 & 69.9    & 56.4 & 56.5 & 69.0 \\
    ROMTrack-256\cite{ROMTrack} & ICCV23    & 69.3 & 78.8 & 75.6    & 48.9 & 59.3 & 55.0    & 72.9 & 82.9 & 70.2    & - & - & 69.7 \\
    ARTrack-256\cite{ARTrack} & CVPR23      & \textbf{\color{blue}{70.4}} & 79.5 & \textbf{\color{blue}{76.6}}    & 46.4 & 56.5 & 52.3    & 73.5 & 82.2 & 70.9    & \textbf{\color{blue}{57.5}} & - & 67.7 \\
    SeqTrack-B256\cite{seqtrack} & CVPR23   & 69.9 & \textbf{\color{blue}{79.7}} & 76.3    & \textbf{\color{blue}{49.5}} & \textbf{\color{blue}{60.8}} & \textbf{\color{blue}{56.3}}    & \textbf{\color{red}{74.7}} & \textbf{\color{red}{84.7}} & \textbf{\color{red}{71.8}}    & 54.9 & - & 69.2 \\
    OSTrack-256\cite{ostrack} &ECCV22       & 69.1 & 78.7 & 75.2    & 47.4 & 57.3 & 53.3    & 71.0 & 80.4 & 68.2    & 54.3 & -   & 68.3\\
    VideoTrack\cite{VideoTrack1}  & CVPR23   & 70.2 & - & 76.4       & -    & -    & -       & 72.9 & 81.9 & 69.8    & - & - & 69.7 \\
    SimTrack-B/16\cite{simtrack}&ECCV22     & 69.3 & 78.5 &-        &-     &-     &-        & 68.6 & 78.9 & 62.4    & 54.8 & 53.8 & 69.8 \\ 
    MixFormer-22k\cite{mixformer}&CVPR22    & 69.2 & 78.7 & 74.7    & -    & -    & -       & 70.7 & 80.0 & 67.8    & - & - & 70.4 \\
    AiATrack-320\cite{aiatrack}&ECCV22          & 69.0 & 79.4 & 73.8    &-     &-     &-        & 69.6 & 80.0 & 63.2    & - & - & \textbf{\color{blue}{70.6}} \\ 
    STARK\cite{stark} & ICCV21              & 67.1 & 77.0 & -       & -    &-     & -       & 68.8 & 78.1 & 64.1    & - & - &-     \\
    AutoMatch\cite{AutoMatch}&ICCV21        & 58.3 &-     & 59.9    & -    & -    &-        & 65.2 & 76.6 & 54.3    & 47.2 & 43.5     & - \\
    TransT \cite{transt}& CVPR21            & 64.9 & 73.8 & 69.0    & -    & -    & -       & 67.1 & 76.8 & 60.9    & 50.7 & 51.7 & 69.1 \\
    TrDiMP\cite{trdimp} & CVPR21            & 63.9 &  -   & 61.4    &-     &-     & -       & 67.1 & 77.7 & 58.3    & - & - & 67.5 \\
    Ocean \cite{Ocean}&  ECCV 20            & 56.0 & 65.1 & 56.6    &-     & -    &-        & 61.1 & 72.1 & 47.3    & 38.4    & 37.7   &-     \\
    SiamPRN++\cite{SiamRPN++}&CVPR19        & 49.6 & 56.9 & 49.1    & 34.0 & 41.6 & 39.6    & 51.7 & 61.6 & 32.5    & 41.3 & 41.2 & 61.0 \\
    ECO \cite{ECO} & ICCV 17                & 32.4 & 33.8 & 30.1    & 22.0 & 25.2 & 24.0    & 31.6 & 30.9 & 11.1    & 32.6 & 31.7 & 53.5 \\
    MDNet \cite{MDNET} & CVPR16             & 39.7 & 46.0 & 37.3    & 27.9 & 34.9 & 31.8    & 29.9 & 30.3 & 9.9     & 38.0 & 37.1 & 52.8 \\
    SiamFC \cite{SiamFC} & ECCVW16          & 33.6 & 42.0 & 33.9    & 23.0 & 31.1 & 26.9    & 34.8 & 35.3 & 9.8     & 29.5 & 28.6 & 46.8 \\
    \midrule
    
    \multicolumn{14}{l}{\multirow{1}{*}{\textit{Some Trackers with Higher Resolution}} }\\
    \midrule
    OSTrack-384\cite{ostrack}&ECCV22        & 71.1 & 81.1 & 77.6    & 50.5 & 61.3 & 57.6    & 73.7 & 83.2 & 70.8    & 55.9 &-  &70.7  \\
    SeqTrack-B384\cite{seqtrack} & CVPR23   & 71.5 & 81.1 & 77.8    & 50.5 & 61.6 & 57.5    & 74.5 & \textbf{\color{blue}{84.3}} & 71.4    & 56.4 & - & 68.6 \\
    ARTrack-384\cite{ARTrack} & CVPR23      & \textbf{\color{blue}{72.6}} & \textbf{\color{blue}{81.7}} & \textbf{\color{blue}{79.1}}    & \textbf{\color{blue}{51.9}} & 62.0 & 58.5    & \textbf{\color{blue}{75.5}} & \textbf{\color{blue}{84.3}} & \textbf{\color{blue}{74.3}}    & \textbf{\color{red}{59.8}} & - & 70.5 \\
    ROMTrack-384\cite{ROMTrack} & ICCV23    & 71.4 & 81.4 & 78.2    & 51.3 & \textbf{\color{blue}{62.4}} & \textbf{\color{blue}{58.6}}    & 74.2 & \textbf{\color{blue}{84.3}} & 72.4    & - & - & 70.5 \\
    F-BDMTrack-384\cite{F-BDMTrack}&ICCV23  & 72.0 & 81.5 & 77.7    & 50.8 & 61.3 & 57.8    & 75.4 & \textbf{\color{blue}{84.3}} & 72.9    & 57.8 & \textbf{\color{blue}{59.4}} & \textbf{\color{blue}{70.9}}\\
    \midrule
    {\mytracker}-384 & Ours                      & \textbf{\color{red}{72.7}} & \textbf{\color{red}{82.9}} & \textbf{\color{red}{80.2}}    & \textbf{\color{red}{52.7}} & \textbf{\color{red}{64.2}} & \textbf{\color{red}{60.8}}    & \textbf{\color{red}{76.0}} & \textbf{\color{red}{85.2}} & \textbf{\color{red}{74.9}}    & \textbf{\color{blue}{59.3}}  & \textbf{\color{red}{62.3}}  &\textbf{\color{red}{71.2}}\\
    \bottomrule    
    \end{tabular}
    }

\label{tab:1}
\end{table*}

\begin{table*}[t]
\caption{ Performance comparisons with state-of-the-art trackers on the test set of TrackingNet. The top two results are highlighted with {\color{red}red} and {\color{blue}blue} fonts respectively.}
\centering
\resizebox{\textwidth}{!}{
    \fontsize{9}{12}\selectfont
    \begin{tabular}{c|ccccccccccc|cc}
    \toprule
    &SiamFC &ECO &SiamRPN++ &TransT &STARK &MixFormer-22k &AiATrack &OSTrack &ARTrack &SeqTrack &F-BDMTrack &{\mytracker}-256 &{\mytracker}-384\\
    &\cite{SiamFC}& \cite{ECO}&\cite{SiamRPN++} &\cite{transt} &\cite{stark} & \cite{mixformer} & \cite{aiatrack} & \cite{ostrack} &\cite{ARTrack} &\cite{seqtrack} &\cite{F-BDMTrack} &ours & ours\\
    \cline{1-14}
    AUC(\%)           &57.1 &55.4 &73.3 &81.4 &82.0 &83.1 &82.7 &83.9 &\textbf{\color{red}85.1} &83.9 &83.7 &83.8 & \textbf{\color{blue}84.8} \\
    $\rm P_{Norm}$(\%)&66.3 &61.8 &80.0 &86.7 &86.9 &88.1 &87.8 &88.5 &\textbf{\color{blue}89.1} &88.8 &88.3 &88.6 & \textbf{\color{red}89.3}\\
    P(\%)             &53.3 &49.2 &69.4 &80.3 &-    &81.6 &80.4 &83.2 &\textbf{\color{red}84.8} &83.6 &82.6 &83.1 & \textbf{\color{blue}84.3}\\
    \bottomrule 
    \end{tabular}}

\label{tab:2}
\end{table*}

\subsection{Results and Comparisons}
To demonstrate the effectiveness of our method, \mytracker is compared with the current SOTA trackers on six datasets, i.e., LaSOT \cite{lasot}, LaSOT$_{ext}$ \cite{lasot-ext}, GOT-10K \cite{got10k}, TNL2K \cite{tnl2k}, UAV123 \cite{uav123}, TrackingNet \cite{trackingnet}.

\textbf{LaSOT} \cite{lasot}. LaSOT consists of 280 videos for testing, serving as a challenging large-scale long-term tracking benchmark. As illustrated in \cref{tab:1}, we compare the results of the {\mytracker} with the previous SOTA trackers. 
The results show that \mytracker-256 achieved 71.4\% AUC on LaSOT \cite{lasot}, significantly outperforming other trackers at the same resolution.
And \mytracker-384 with an AUC score of 72.7\% surpassing all other trackers without bells and whistles. 
As shown at \cref{fig:zhizhuwang}, {\mytracker} demonstrate competitive performance in all challenges on LaSOT.
Particularly, \cref{fig:motion} demonstrates {\mytracker} outperforms previous SOTA trackers in camera motion and motion blur challenges.
The outstanding performance on the LaSOT benchmark mentioned above proves the effectiveness of our approach in mining spatio-temporal information.

\textbf{LaSOT$_{ext}$} \cite{lasot-ext}. LaSOT$_{ext}$ expands upon the LaSOT \cite{lasot} dataset by adding 150 videos. These introduced sequences pose significant challenges due to the presence of numerous interference of similar objects in the videos. The results presented in \cref{tab:1} demonstrate that our \mytracker-256 outperforms all other trackers by a substantial margin, achieving the highest P$_{norm}$ score of 62.2\% and surpassing ARTrack \cite{ARTrack} by 1.4\%. Additionally, our higher resolution model \mytracker-384 also significantly outperforms previous trackers in all three metrics. This establishes a new state-of-the-art on LaSOT$_{ext}$, indicating that our tracker has a robust discrimination ability against similar distractors.

\textbf{GOT-10k} \cite{got10k}. In the GOT-10k dataset's test set, a one-shot tracking rule is applied. This rule mandates that trackers are solely trained on the GOT-10k training split, and there is no overlap in object classes between the train and test splits. We adhere to this protocol for training our model and evaluate the results by submitting them to the official evaluation server. As indicated in \cref{tab:1}, \mytracker-384 and \mytracker-256 outperform most of the previous trackers. This highlights our trackers' ability to perceive the target's changing state. 
The exceptional performance on this one-shot benchmark underscores the effectiveness of {\mytracker} in extracting spatio-temporal information for unseen classes in an adaptive and autoregression manner.

\textbf{TNL2K} \cite{tnl2k}. 
TNL2K is a tracking dataset consisting of 700 videos available for testing. \mytracker-256 achieved a new state-of-the-art score of 57.8\% in AUC at the same resolution, as shown in the \cref{tab:1}. \mytracker-384 also demonstrated outstanding performance, with AUC score of 59.3\%.

\textbf{UAV123} \cite{uav123}.
UAV123 is a dataset for unmanned aerial vehicles, comprising 123 videos. In the \cref{tab:1}, we present the results, including our model and previous state-of-the-art trackers such as OSTrack, MixFormer, TransT, and STARK, among others. Our model significantly outperforms these methods, achieving two AUC scores of 70.7\% and 71.2\% with a considerable margin.

\textbf{TrackingNet} \cite{trackingnet}. 
The TrackingNet benchmark comprises 511 testing sequences. As shown in \cref{tab:2}, \mytracker-384 demonstrated competitive performance compared to previous state-of-the-art trackers. {\mytracker} gets the best P$_{norm}$, surpassing the previous best-performing tracker ARtrack \cite{ARTrack}.

\begin{table}
    \caption{ Ablation study for important components. Blank denotes the component is used by default, while \ding{55} represents the component is removed. Performance is evaluated on LaSOT\cite{lasot}.}
    \centering
    \resizebox{\linewidth}{!}{
    \fontsize{9}{12}\selectfont
    \begin{tabular}{c|ccccc|ccc}
    \toprule
     \# & Decoder & STM & TA & Q$_{all}$ & Autoregression & AUC(\%) & P$_{norm}$(\%) & P(\%)\\
     \midrule
     1 & \ding{55} & \ding{55} & & &  &70.5 & 80.7 &77.5 \\
     2 & & & \ding{55} & \ding{55}  &  & 70.8 & 80.9 & 77.6\\ 
     3 & & & & \ding{55} & \ding{55} & 70.5 & 80.6 & 77.4     \\
     4 & & & & &    & \textbf{71.4} & \textbf{81.9} & \textbf{78.6}\\ 
    \bottomrule 
    \end{tabular}
        }
    
    \label{tab:ablation}
\end{table} 

\subsection{Ablation Study and Analysis}
To demonstrate the effectiveness of our proposed spatio-temporal information learning method, we conducted some ablation study using {\mytracker}-256 as the baseline on LaSOT \cite{lasot}.

\textbf{Spatial encoder}.
Recently, the vanilla ViT-Base \cite{vit} model pretrained with MAE \cite{MAE}, has been frequently used as a backbone for feature extraction and fusion. For a fair comparison, we conducted a study using ViT as the spatial encoder in {\mytracker}-256, and the results as shown in \cref{tab:encoder}. 
The ViT-based {\mytracker} improved by 0.5\% in terms of success compared to only ViT, which indicates the effectiveness of the proposed temporal decoder with learnable and autoregressive queries.
To avoid damaging the integrity of spatial information, we use a hierarchical transformer (HiViT \cite{hivit}) as the encoder.
The HiViT-based {\mytracker} improved by 0.9\% compared to only HiViT.
In summary, the HiViT-based {\mytracker}-256 outperforms the ViT-based {\mytracker}-256 by 0.8\% in AUC. 
It also shows that our spatial encoder can better capture the spatial features of targets.
This is attributed to the fact that, unlike ViT where images are downsampled 16 times simultaneously, HiViT progressively downsamples images before entering the attention block. 
So it facilitates the effective preservation of spatial features.

\begin{table}[t]
    \caption{ The effectiveness of our method using other encoders on LaSOT\cite{lasot}.}
    \centering
        \fontsize{8}{9}\selectfont
        \begin{tabular}{c|c|ccc}
        \toprule
         Encoder & Temporal Decoder &AUC(\%) & P$_{norm}$(\%) &P(\%)\\
         \midrule
         \multicolumn{1}{c|}{\multirow{2}{*}{ViT\cite{vit}} } & - & 69.1 & 79.1 & 75.1\\
          & \checkmark & 69.6 & 79.5 & 75.6\\ 
         \midrule
         \multicolumn{1}{c|}{\multirow{2}{*}{HiViT\cite{hivit}} } & - & 70.5 & 80.7 & 77.5 \\
         & \checkmark & \textbf{71.4} & \textbf{81.9} & \textbf{78.6} \\
        \bottomrule 
        \end{tabular}
    \label{tab:encoder}
\end{table} 

\begin{table}[t]
    \caption{ Experimental studies on the number of decoder layers.}
    \centering
        \fontsize{8}{11}\selectfont
        \begin{tabular}{c|ccc}
        \toprule
         Number of Decoder Layers(M) & AUC(\%) & P$_{norm}$(\%) &P(\%)\\
         \midrule
         1 & 70.5 & 80.6 & 77.2\\
         3 & \textbf{71.4} & \textbf{81.9} & \textbf{78.6} \\
         6 & 71.1 & 81.2 & 77.9 \\
        \bottomrule 
        \end{tabular}
    
    \label{tab:decoder}
\end{table} 


\textbf{Temporal decoder}.
The temporal decoder is a key module for instantaneous spatio-temporal information learning. To demonstrate the effectiveness of this module, we tested the impact of different numbers of temporal decoder layers ($M$) on our method. 
When $M$ is 0, it is equivalent to removing the temporal decoder and STM from our network, leaving a spatial encoder, degenerated into a structure similar to OSTrack \cite{ostrack}.
As demonstrated in the \cref{tab:ablation} (\#1), the tracker only has a spatial encoder and achieved an AUC score of 70.5\% on the LaSOT benchmark, which is 0.9\% lower than the {\mytracker} that uses both spatio-temporal information (as shown in \cref{tab:ablation} (\#4)).
This proves that learning solely based on spatial features cannot accurately track the target.
As illustrated in \cref{tab:decoder}, when the number of decoder layers increases from 1 to 3, the decoder learns richer temporal information, resulting in better performance. 
The three variants in \cref{tab:decoder} are trained using the same configuration, except for the number of layers. 
When the number of layers $M$ is 6, {\mytracker}'s performance slightly decreases.

\textbf{Temporal queries and length of spatio-temporal information}.
We introduce a target query to describe the state of the target. The input of the decoder is the target query of the previous $m$-1 frames and the current frame, which means that the current frame's state references the previous frames.
So the length of spatio-temporal information obtained by the model is $m$.
We investigate the impact of different spatio-temporal information lengths on {\mytracker} performance.
As shown in \cref{tab:frame} (\#1, \#2, \#3), the performance of the model increases as the spatio-temporal range($m$) increases from 1 to 4.
This proves that when there are more frames for model reference, more target state changes can be obtained, thereby achieving better performance.
However, when $m$ is 8 and 16, the performance of the model slightly decreases. This is because the number of video sequences($n$) used in each iteration is too small, resulting in a lack of generalization ability.
\begin{table}[t]
    \caption{ Influence of the length of spatio-temporal information.}
    \centering
        \fontsize{9}{11}\selectfont
        \begin{tabular}{c|c|c|ccc}
        \toprule
         Batchsize & m & n & AUC(\%) & P$_{norm}$(\%) & P(\%)\\
         \midrule
         32 & 1  & 32 & 70.5 & 80.6 & 77.4 \\
         32 & 2  & 16 & 70.8 & 80.8 & 77.5 \\
         32 & 4  & 8  & \textbf{71.4} & \textbf{81.9} &  \textbf{78.6} \\
         32 & 8  & 4  & 70.6 & 80.7 & 77.7  \\
         32 & 16 & 2  & 70.2 & 80.5 & 77.3  \\
        \bottomrule 
        \end{tabular}
    
    \label{tab:frame}
\end{table}

\textbf{Temporal attention (TA)}. To demonstrate temporal attention's key role in collecting target state change information across frames, we conducted experiments using self-attention as a substitute.
As a result, the AUC, P$_{norm}$, and P obtained by the self-attention-based tracker on LaSOT \cite{lasot} were 70.7\%, 80.9\%, and 77.8\%, respectively, which are lower than tracker 0.9\% on AUC. This experiment proves that our designed temporal attention is more effective in capturing the target motion trend across frames.

\textbf{Spatial and temporal fusion operation module(STM)}.
In \mytracker, STM is a simple and effective module that does not require learning parameter weights.
It can enhance important regions using simple arithmetic operations.
We attempted to use a self-attention layer (i.e., one block of ViT \cite{vit}) to fuse temporal information and spatial features, and the result was a 70.5\% (-0.9\% compared to STM) AUC score on the LaSOT benchmark.

\begin{figure}[t]
    \centering
    \includegraphics[height=4.5cm]{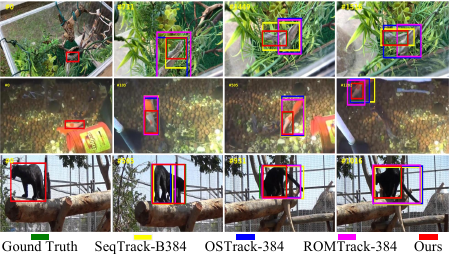}
    \caption{Comparison tracking result with other three SOTA trackers on LaSOT\cite{lasot} benchmark.}
    \label{fig:vis_box}
\end{figure} 
\begin{figure}[t]
    \centering
    \includegraphics[height=5.0cm]{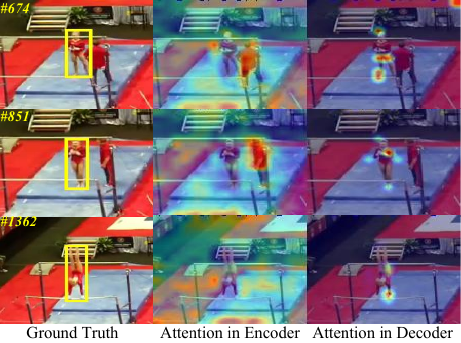}
    \caption{The comparison attention map in the spatial encoder and the temporal decoder on LaSOT\cite{lasot}. The first column is the ground truth, the second column is the self-attention in the encoder, and the third column is the multi-head attention(MHA) in the decoder.}
    \label{fig:vis_att}
\end{figure} 

\textbf{Visualization and qualitative comparison.} 
To intuitively demonstrate the effectiveness of our proposed continuous spatio-temporal modeling approach in occluded scenes, we visualized the tracking results of {\mytracker} and previous 
three SOTA models (i.e., SeqTrack \cite{seqtrack}, OSTrack \cite{ostrack}, ROMTrack \cite{ROMTrack}). 
As shown in the \cref{fig:vis_box}, due to our tracker's excellent continuous spatio-temporal modeling ability to capture more delicate and subtle changes in the target state, our tracker achieved the most accurate tracking in these occlusion sequences. 
In addition, to demonstrate the effectiveness of temporal information, we also visualized the attention maps of the spatial encoder and temporal decoder, as demonstrated in \cref{fig:vis_att}. 
It can be seen that under the guidance of temporal information, the model pays more attention to the location of the target.

\section{Conclusion }
We propose {\mytracker} for continuous spatio-tmeporal information modeling from a novel perspective. 
We design a temporal decoder with temporal attention and temporal queries, which captures the motion trend to discriminate the target from similar objects.
Empirical evaluation shows that our method is effective and achieves competitive performance compared to previous SOTA trackers.

\textit{Limitation}.
Our proposed method uses learnable and autoregressive target queries to capture the concentrated spatio-temporal information, reducing the unnecessary background interference introduced by some methods such as dynamic templates. 
So we believe that {\mytracker} with longer spatio-temporal information can exhibit stronger performance.
However, due to the limitations of GPU memory, the modeling of more long-term spatio-temporal information in {\mytracker} has not been thoroughly explored.

\textit{Acknowledgement.}
This work is supported by the Project of Guangxi Science and Technology (No.2022GXNSFDA035079), the National Natural Science Foundation of China (No.U23A20383 and U21A20474), the Guangxi ``Young Bagui Scholar" Teams for Innovation and Research Project, the Guangxi ``Bagui Scholar" Teams for Innovation and Research Project, the Guangxi Collaborative Innovation Center of Multi-source Information Integration and Intelligent Processing, and the Guangxi Talent Highland Project of Big Data Intelligence and Application.
{
    \small
    \bibliographystyle{ieeenat_fullname}
    \bibliography{main}

\begin{thebibliography}{56}
\providecommand{\natexlab}[1]{#1}
\providecommand{\url}[1]{\texttt{#1}}
\expandafter\ifx\csname urlstyle\endcsname\relax
  \providecommand{\doi}[1]{doi: #1}\else
  \providecommand{\doi}{doi: \begingroup \urlstyle{rm}\Url}\fi

\bibitem[Bertinetto et~al.(2016)Bertinetto, Valmadre, Henriques, Vedaldi, and Torr]{SiamFC}
Luca Bertinetto, Jack Valmadre, Jo{\~{a}}o~F. Henriques, Andrea Vedaldi, and Philip H.~S. Torr.
\newblock Fully-convolutional siamese networks for object tracking.
\newblock In \emph{{ECCV} Workshops}, pages 850--865, 2016.

\bibitem[Bhat et~al.(2020)Bhat, Danelljan, Gool, and Timofte]{KYS}
Goutam Bhat, Martin Danelljan, Luc~Van Gool, and Radu Timofte.
\newblock Know your surroundings: Exploiting scene information for object tracking.
\newblock In \emph{{ECCV}}, pages 205--221, 2020.

\bibitem[Cai et~al.(2023)Cai, Liu, Tang, and Wu]{ROMTrack}
Yidong Cai, Jie Liu, Jie Tang, and Gangshan Wu.
\newblock Robust object modeling for visual tracking.
\newblock \emph{CoRR}, abs/2308.05140, 2023.

\bibitem[Cao et~al.(2022)Cao, Huang, Pan, Zhang, Liu, and Fu]{TCTrack}
Ziang Cao, Ziyuan Huang, Liang Pan, Shiwei Zhang, Ziwei Liu, and Changhong Fu.
\newblock Tctrack: Temporal contexts for aerial tracking.
\newblock In \emph{{CVPR}}, pages 14778--14788, 2022.

\bibitem[Cao et~al.(2023)Cao, Huang, Pan, Zhang, Liu, and Fu]{TCTrack++}
Ziang Cao, Ziyuan Huang, Liang Pan, Shiwei Zhang, Ziwei Liu, and Changhong Fu.
\newblock Towards real-world visual tracking with temporal contexts.
\newblock \emph{IEEE Transactions on Pattern Analysis and Machine Intelligence}, 2023.

\bibitem[Carion et~al.(2020)Carion, Massa, Synnaeve, Usunier, Kirillov, and Zagoruyko]{DETR}
Nicolas Carion, Francisco Massa, Gabriel Synnaeve, Nicolas Usunier, Alexander Kirillov, and Sergey Zagoruyko.
\newblock End-to-end object detection with transformers.
\newblock In \emph{{ECCV}}, pages 213--229, 2020.

\bibitem[Chen et~al.(2022)Chen, Li, Bai, Qiao, Shen, Li, Gan, Wu, and Ouyang]{simtrack}
Boyu Chen, Peixia Li, Lei Bai, Lei Qiao, Qiuhong Shen, Bo Li, Weihao Gan, Wei Wu, and Wanli Ouyang.
\newblock Backbone is all your need: {A} simplified architecture for visual object tracking.
\newblock In \emph{{ECCV} {(22)}}, pages 375--392, 2022.

\bibitem[Chen et~al.(2021)Chen, Yan, Zhu, Wang, Yang, and Lu]{transt}
Xin Chen, Bin Yan, Jiawen Zhu, Dong Wang, Xiaoyun Yang, and Huchuan Lu.
\newblock Transformer tracking.
\newblock In \emph{{CVPR}}, pages 8126--8135, 2021.

\bibitem[Chen et~al.(2023)Chen, Peng, Wang, Lu, and Hu]{seqtrack}
Xin Chen, Houwen Peng, Dong Wang, Huchuan Lu, and Han Hu.
\newblock Seqtrack: Sequence to sequence learning for visual object tracking.
\newblock In \emph{CVPR}, 2023.

\bibitem[Cheng et~al.(2022)Cheng, Wang, and Li]{surveillance0}
Linsong Cheng, Jiliang Wang, and Yinghui Li.
\newblock Vitrack: Efficient tracking on the edge for commodity video surveillance systems.
\newblock \emph{IEEE Transactions on Parallel and Distributed Systems}, 33\penalty0 (3):\penalty0 723--735, 2022.

\bibitem[Cui et~al.(2022)Cui, Jiang, Wang, and Wu]{mixformer}
Yutao Cui, Cheng Jiang, Limin Wang, and Gangshan Wu.
\newblock Mixformer: End-to-end tracking with iterative mixed attention.
\newblock In \emph{{CVPR}}, pages 13598--13608, 2022.

\bibitem[Danelljan et~al.(2017)Danelljan, Bhat, Khan, and Felsberg]{ECO}
Martin Danelljan, Goutam Bhat, Fahad~Shahbaz Khan, and Michael Felsberg.
\newblock {ECO:} efficient convolution operators for tracking.
\newblock In \emph{{CVPR}}, pages 6931--6939, 2017.

\bibitem[Dosovitskiy et~al.(2021)Dosovitskiy, Beyer, Kolesnikov, Weissenborn, Zhai, Unterthiner, Dehghani, Minderer, Heigold, Gelly, Uszkoreit, and Houlsby]{vit}
Alexey Dosovitskiy, Lucas Beyer, Alexander Kolesnikov, Dirk Weissenborn, Xiaohua Zhai, Thomas Unterthiner, Mostafa Dehghani, Matthias Minderer, Georg Heigold, Sylvain Gelly, Jakob Uszkoreit, and Neil Houlsby.
\newblock An image is worth 16x16 words: Transformers for image recognition at scale.
\newblock In \emph{{ICLR}}, 2021.

\bibitem[Ettinger et~al.(2021)Ettinger, Cheng, Caine, Liu, Zhao, Pradhan, Chai, Sapp, Qi, Zhou, Yang, Chouard, Sun, Ngiam, Vasudevan, McCauley, Shlens, and Anguelov]{autodriving1}
Scott Ettinger, Shuyang Cheng, Benjamin Caine, Chenxi Liu, Hang Zhao, Sabeek Pradhan, Yuning Chai, Ben Sapp, Charles~R. Qi, Yin Zhou, Zoey Yang, Aur\'elien Chouard, Pei Sun, Jiquan Ngiam, Vijay Vasudevan, Alexander McCauley, Jonathon Shlens, and Dragomir Anguelov.
\newblock Large scale interactive motion forecasting for autonomous driving: The waymo open motion dataset.
\newblock In \emph{Proceedings of the IEEE/CVF International Conference on Computer Vision (ICCV)}, pages 9710--9719, 2021.

\bibitem[Fan et~al.(2019)Fan, Lin, Yang, Chu, Deng, Yu, Bai, Xu, Liao, and Ling]{lasot}
Heng Fan, Liting Lin, Fan Yang, Peng Chu, Ge Deng, Sijia Yu, Hexin Bai, Yong Xu, Chunyuan Liao, and Haibin Ling.
\newblock Lasot: {A} high-quality benchmark for large-scale single object tracking.
\newblock In \emph{{CVPR}}, pages 5374--5383, 2019.

\bibitem[Fan et~al.(2021)Fan, Bai, Lin, Yang, Chu, Deng, Yu, Harshit, Huang, Liu, Xu, Liao, Yuan, and Ling]{lasot-ext}
Heng Fan, Hexin Bai, Liting Lin, Fan Yang, Peng Chu, Ge Deng, Sijia Yu, Harshit, Mingzhen Huang, Juehuan Liu, Yong Xu, Chunyuan Liao, Lin Yuan, and Haibin Ling.
\newblock Lasot: {A} high-quality large-scale single object tracking benchmark.
\newblock \emph{Int. J. Comput. Vis.}, pages 439--461, 2021.

\bibitem[Fu et~al.(2021)Fu, Liu, Fu, and Wang]{STMTrack}
Zhihong Fu, Qingjie Liu, Zehua Fu, and Yunhong Wang.
\newblock Stmtrack: Template-free visual tracking with space-time memory networks.
\newblock In \emph{{CVPR}}, pages 13774--13783, 2021.

\bibitem[Fu et~al.(2022)Fu, Fu, Liu, Cai, and Wang]{sparseTT}
Zhihong Fu, Zehua Fu, Qingjie Liu, Wenrui Cai, and Yunhong Wang.
\newblock Sparsett: Visual tracking with sparse transformers.
\newblock \emph{arXiv preprint arXiv:2205.03776}, 2022.

\bibitem[Gao et~al.(2022)Gao, Zhou, Ma, Wang, and Yuan]{aiatrack}
Shenyuan Gao, Chunluan Zhou, Chao Ma, Xinggang Wang, and Junsong Yuan.
\newblock Aiatrack: Attention in attention for transformer visual tracking.
\newblock In \emph{{ECCV} {(22)}}, pages 146--164, 2022.

\bibitem[Gao et~al.(2023)Gao, Zhou, and Zhang]{GRM}
Shenyuan Gao, Chunluan Zhou, and Jun Zhang.
\newblock Generalized relation modeling for transformer tracking.
\newblock In \emph{Proceedings of the IEEE/CVF Conference on Computer Vision and Pattern Recognition}, pages 18686--18695, 2023.

\bibitem[He et~al.(2016)He, Zhang, Ren, and Sun]{resnet1}
Kaiming He, Xiangyu Zhang, Shaoqing Ren, and Jian Sun.
\newblock Deep residual learning for image recognition.
\newblock In \emph{Proceedings of the IEEE Conference on Computer Vision and Pattern Recognition (CVPR)}, 2016.

\bibitem[He et~al.(2022)He, Chen, Xie, Li, Doll{\'{a}}r, and Girshick]{MAE}
Kaiming He, Xinlei Chen, Saining Xie, Yanghao Li, Piotr Doll{\'{a}}r, and Ross~B. Girshick.
\newblock Masked autoencoders are scalable vision learners.
\newblock In \emph{{CVPR}}, pages 15979--15988, 2022.

\bibitem[Huang et~al.(2021)Huang, Zhao, and Huang]{got10k}
Lianghua Huang, Xin Zhao, and Kaiqi Huang.
\newblock Got-10k: {A} large high-diversity benchmark for generic object tracking in the wild.
\newblock \emph{{IEEE} Trans. Pattern Anal. Mach. Intell.}, 43\penalty0 (5):\penalty0 1562--1577, 2021.

\bibitem[Krizhevsky et~al.(2012)Krizhevsky, Sutskever, and Hinton]{alexnet}
Alex Krizhevsky, Ilya Sutskever, and Geoffrey~E. Hinton.
\newblock Imagenet classification with deep convolutional neural networks.
\newblock In \emph{{NIPS}}, pages 1106--1114, 2012.

\bibitem[Li et~al.(2019)Li, Wu, Wang, Zhang, Xing, and Yan]{SiamRPN++}
Bo Li, Wei Wu, Qiang Wang, Fangyi Zhang, Junliang Xing, and Junjie Yan.
\newblock Siamrpn++: Evolution of siamese visual tracking with very deep networks.
\newblock In \emph{{CVPR}}, pages 4282--4291, 2019.

\bibitem[Lin et~al.(2022)Lin, Fan, Zhang, Xu, and Ling]{SwinTrack}
Liting Lin, Heng Fan, Zhipeng Zhang, Yong Xu, and Haibin Ling.
\newblock Swintrack: A simple and strong baseline for transformer tracking.
\newblock \emph{Advances in Neural Information Processing Systems}, 35:\penalty0 16743--16754, 2022.

\bibitem[Lin et~al.(2014)Lin, Maire, Belongie, Hays, Perona, Ramanan, Doll{\'{a}}r, and Zitnick]{coco}
Tsung{-}Yi Lin, Michael Maire, Serge~J. Belongie, James Hays, Pietro Perona, Deva Ramanan, Piotr Doll{\'{a}}r, and C.~Lawrence Zitnick.
\newblock Microsoft {COCO:} common objects in context.
\newblock In \emph{{ECCV}}, pages 740--755, 2014.

\bibitem[Lin et~al.(2017)Lin, Goyal, Girshick, He, and Doll{\'{a}}r]{focalloss}
Tsung{-}Yi Lin, Priya Goyal, Ross~B. Girshick, Kaiming He, and Piotr Doll{\'{a}}r.
\newblock Focal loss for dense object detection.
\newblock In \emph{{ICCV}}, pages 2999--3007, 2017.

\bibitem[Loshchilov and Hutter(2019)]{adamw}
Ilya Loshchilov and Frank Hutter.
\newblock Decoupled weight decay regularization.
\newblock In \emph{{ICLR}}, 2019.

\bibitem[Mayer et~al.(2022)Mayer, Danelljan, Bhat, Paul, Paudel, Yu, and Van~Gool]{ToMP}
Christoph Mayer, Martin Danelljan, Goutam Bhat, Matthieu Paul, Danda~Pani Paudel, Fisher Yu, and Luc Van~Gool.
\newblock Transforming model prediction for tracking.
\newblock In \emph{Proceedings of the IEEE/CVF conference on computer vision and pattern recognition}, pages 8731--8740, 2022.

\bibitem[Meinhardt et~al.(2022)Meinhardt, Kirillov, Leal-Taixe, and Feichtenhofer]{TrackFormer}
Tim Meinhardt, Alexander Kirillov, Laura Leal-Taixe, and Christoph Feichtenhofer.
\newblock Trackformer: Multi-object tracking with transformers.
\newblock In \emph{The IEEE Conference on Computer Vision and Pattern Recognition (CVPR)}, 2022.

\bibitem[Mueller et~al.(2016)Mueller, Smith, and Ghanem]{uav123}
Matthias Mueller, Neil Smith, and Bernard Ghanem.
\newblock A benchmark and simulator for {UAV} tracking.
\newblock In \emph{{ECCV}}, pages 445--461, 2016.

\bibitem[M{\"{u}}ller et~al.(2018)M{\"{u}}ller, Bibi, Giancola, Al{-}Subaihi, and Ghanem]{trackingnet}
Matthias M{\"{u}}ller, Adel Bibi, Silvio Giancola, Salman Al{-}Subaihi, and Bernard Ghanem.
\newblock Trackingnet: {A} large-scale dataset and benchmark for object tracking in the wild.
\newblock In \emph{{ECCV}}, pages 310--327, 2018.

\bibitem[Nam and Han(2016)]{MDNET}
Hyeonseob Nam and Bohyung Han.
\newblock Learning multi-domain convolutional neural networks for visual tracking.
\newblock In \emph{{CVPR}}, pages 4293--4302, 2016.

\bibitem[Premachandra et~al.(2020)Premachandra, Ueda, and Suzuki]{autodriving0}
Chinthaka Premachandra, Shohei Ueda, and Yuya Suzuki.
\newblock Detection and tracking of moving objects at road intersections using a 360-degree camera for driver assistance and automated driving.
\newblock \emph{IEEE Access}, 8:\penalty0 135652--135660, 2020.

\bibitem[Rezatofighi et~al.(2019)Rezatofighi, Tsoi, Gwak, Sadeghian, Reid, and Savarese]{giouloss}
Hamid Rezatofighi, Nathan Tsoi, JunYoung Gwak, Amir Sadeghian, Ian~D. Reid, and Silvio Savarese.
\newblock Generalized intersection over union: {A} metric and a loss for bounding box regression.
\newblock In \emph{{IEEE} Conference on Computer Vision and Pattern Recognition, {CVPR} 2019, Long Beach, CA, USA, June 16-20, 2019}, pages 658--666. Computer Vision Foundation / {IEEE}, 2019.

\bibitem[Shehzed et~al.(2019)Shehzed, Jalal, and Kim]{surveillance1}
Ahsan Shehzed, Ahmad Jalal, and Kibum Kim.
\newblock Multi-person tracking in smart surveillance system for crowd counting and normal/abnormal events detection.
\newblock In \emph{2019 International Conference on Applied and Engineering Mathematics (ICAEM)}, pages 163--168, 2019.

\bibitem[Song et~al.(2023)Song, Luo, Yu, Chen, and Yang]{CTTrack}
Zikai Song, Run Luo, Junqing Yu, Yi-Ping~Phoebe Chen, and Wei Yang.
\newblock Compact transformer tracker with correlative masked modeling.
\newblock In \emph{Proceedings of the AAAI Conference on Artificial Intelligence (AAAI)}, 2023.

\bibitem[Sun et~al.(2020)Sun, Cao, Jiang, Zhang, Xie, Yuan, Wang, and Luo]{TransTrack}
Peize Sun, Jinkun Cao, Yi Jiang, Rufeng Zhang, Enze Xie, Zehuan Yuan, Changhu Wang, and Ping Luo.
\newblock Transtrack: Multiple-object tracking with transformer.
\newblock \emph{arXiv preprint arXiv: 2012.15460}, 2020.

\bibitem[Vaswani et~al.(2017)Vaswani, Shazeer, Parmar, Uszkoreit, Jones, Gomez, Kaiser, and Polosukhin]{transformer}
Ashish Vaswani, Noam Shazeer, Niki Parmar, Jakob Uszkoreit, Llion Jones, Aidan~N Gomez, \L~ukasz Kaiser, and Illia Polosukhin.
\newblock Attention is all you need.
\newblock In \emph{Advances in Neural Information Processing Systems}. Curran Associates, Inc., 2017.

\bibitem[Wang et~al.(2021{\natexlab{a}})Wang, Zhou, Wang, and Li]{trdimp}
Ning Wang, Wengang Zhou, Jie Wang, and Houqiang Li.
\newblock Transformer meets tracker: Exploiting temporal context for robust visual tracking.
\newblock In \emph{CVPR}, pages 1571--1580, 2021{\natexlab{a}}.

\bibitem[Wang et~al.(2021{\natexlab{b}})Wang, Shu, Zhang, Jiang, Wang, Tian, and Wu]{tnl2k}
Xiao Wang, Xiujun Shu, Zhipeng Zhang, Bo Jiang, Yaowei Wang, Yonghong Tian, and Feng Wu.
\newblock Towards more flexible and accurate object tracking with natural language: Algorithms and benchmark.
\newblock In \emph{{CVPR}}, pages 13763--13773, 2021{\natexlab{b}}.

\bibitem[Wang et~al.(2021{\natexlab{c}})Wang, Xu, Wang, Shen, Cheng, Shen, and Xia]{VisTR}
Yuqing Wang, Zhaoliang Xu, Xinlong Wang, Chunhua Shen, Baoshan Cheng, Hao Shen, and Huaxia Xia.
\newblock End-to-end video instance segmentation with transformers.
\newblock In \emph{Proceedings of the IEEE/CVF Conference on Computer Vision and Pattern Recognition (CVPR)}, pages 8741--8750, 2021{\natexlab{c}}.

\bibitem[Wu et~al.(2021)Wu, Jiang, Zhang, Bai, and Bai]{seqformer}
Junfeng Wu, Yi Jiang, Wenqing Zhang, Xiang Bai, and Song Bai.
\newblock Seqformer: a frustratingly simple model for video instance segmentation.
\newblock \emph{arXiv preprint arXiv:2112.08275}, 2021.

\bibitem[Xie et~al.(2022)Xie, Wang, Wang, Cao, Yang, and Zeng]{SBT}
Fei Xie, Chunyu Wang, Guangting Wang, Yue Cao, Wankou Yang, and Wenjun Zeng.
\newblock Correlation-aware deep tracking.
\newblock In \emph{{CVPR}}, pages 8741--8750, 2022.

\bibitem[Xie et~al.(2023)Xie, Chu, Li, Lu, and Ma]{VideoTrack1}
Fei Xie, Lei Chu, Jiahao Li, Yan Lu, and Chao Ma.
\newblock Videotrack: Learning to track objects via video transformer.
\newblock In \emph{Proceedings of the IEEE/CVF Conference on Computer Vision and Pattern Recognition (CVPR)}, pages 22826--22835, 2023.

\bibitem[Xing et~al.(2023)Xing, Yifan, Yongchao, Dahu, and Yihong]{ARTrack}
Wei Xing, Bai Yifan, Zheng Yongchao, Shi Dahu, and Gong Yihong.
\newblock Autoregressive visual tracking.
\newblock In \emph{Proceedings of the IEEE/CVF Conference on Computer Vision and Pattern Recognition (CVPR)}, pages 9697--9706, 2023.

\bibitem[Yan et~al.(2021)Yan, Peng, Fu, Wang, and Lu]{stark}
Bin Yan, Houwen Peng, Jianlong Fu, Dong Wang, and Huchuan Lu.
\newblock Learning spatio-temporal transformer for visual tracking.
\newblock In \emph{ICCV}, pages 10428--10437, 2021.

\bibitem[Yang et~al.(2023)Yang, He, Ma, Yu, and Zhang]{F-BDMTrack}
Dawei Yang, Jianfeng He, Yinchao Ma, Qianjin Yu, and Tianzhu Zhang.
\newblock Foreground-background distribution modeling transformer for visual object tracking.
\newblock In \emph{Proceedings of the IEEE/CVF International Conference on Computer Vision (ICCV)}, pages 10117--10127, 2023.

\bibitem[Yang et~al.(2019)Yang, Fan, and Xu]{VIS1}
Linjie Yang, Yuchen Fan, and Ning Xu.
\newblock Video instance segmentation.
\newblock In \emph{Proceedings of the IEEE/CVF International Conference on Computer Vision (ICCV)}, 2019.

\bibitem[Ye et~al.(2022)Ye, Chang, Ma, Shan, and Chen]{ostrack}
Botao Ye, Hong Chang, Bingpeng Ma, Shiguang Shan, and Xilin Chen.
\newblock Joint feature learning and relation modeling for tracking: {A} one-stream framework.
\newblock In \emph{{ECCV} {(22)}}, pages 341--357, 2022.

\bibitem[Zeng et~al.(2022)Zeng, Dong, Zhang, Wang, Zhang, and Wei]{motr1}
Fangao Zeng, Bin Dong, Yuang Zhang, Tiancai Wang, Xiangyu Zhang, and Yichen Wei.
\newblock {MOTR:} end-to-end multiple-object tracking with transformer.
\newblock In \emph{{ECCV} {(27)}}, pages 659--675, 2022.

\bibitem[Zhang et~al.(2019)Zhang, Gonzalez{-}Garcia, van~de Weijer, Danelljan, and Khan]{updateNet}
Lichao Zhang, Abel Gonzalez{-}Garcia, Joost van~de Weijer, Martin Danelljan, and Fahad~Shahbaz Khan.
\newblock Learning the model update for siamese trackers.
\newblock In \emph{{ICCV}}, pages 4009--4018, 2019.

\bibitem[Zhang et~al.(2023)Zhang, Tian, Xie, Huang, Dai, Ye, and Tian]{hivit}
Xiaosong Zhang, Yunjie Tian, Lingxi Xie, Wei Huang, Qi Dai, Qixiang Ye, and Qi Tian.
\newblock Hivit: A simpler and more efficient design of hierarchical vision transformer.
\newblock In \emph{International Conference on Learning Representations}, 2023.

\bibitem[Zhang et~al.(2020)Zhang, Peng, Fu, Li, and Hu]{Ocean}
Zhipeng Zhang, Houwen Peng, Jianlong Fu, Bing Li, and Weiming Hu.
\newblock Ocean: Object-aware anchor-free tracking.
\newblock In \emph{{ECCV}}, pages 771--787, 2020.

\bibitem[Zhang et~al.(2021)Zhang, Liu, Wang, Li, and Hu]{AutoMatch}
Zhipeng Zhang, Yihao Liu, Xiao Wang, Bing Li, and Weiming Hu.
\newblock Learn to match: Automatic matching network design for visual tracking.
\newblock In \emph{ICCV}, pages 13319--13328, 2021.

\end{thebibliography}
}


\end{document}